\documentclass[letterpaper, twocolumn, 10 pt, conference]{article}  
\usepackage{psfrag}
\usepackage{graphicx}
\usepackage{subfigure}
\usepackage{amsmath}
\usepackage{algorithm} 
\usepackage{algorithmic}
\usepackage{color}
 \usepackage[top=1.5cm, bottom=1.5cm, left=1.5cm, right=1.5cm]{geometry} 

\def\authorrefmark#1{\raisebox{0pt}[0pt][0pt]{\textsuperscript{\footnotesize\ensuremath{\ifcase#1\or *\or \dagger\or \ddagger\or%
    \mathsection\or \mathparagraph\or \|\or **\or \dagger\dagger%
    \or \ddagger\ddagger \else\textsuperscript{\expandafter\romannumeral#1}\fi}}}}

\title{\LARGE \bf
Soft Motion Trajectory Planner for \\ Service Manipulator Robot
}
\date{}
\author{
		 Xavier Broqu\`ere\authorrefmark{1}
		\qquad Daniel Sidobre\authorrefmark{1}
		\qquad Ignacio Herrera-Aguilar\authorrefmark{1}\authorrefmark{3} \\
		\parbox{4 in}{
		\vspace*{3mm} \centering \authorrefmark{1}LAAS - CNRS and Universit\'e de Toulouse \\
	    7, avenue du Colonel Roche \\ 
	    31077 Toulouse, France \\
		\textit{lastname@laas.fr} }
	    \hfill
	    \parbox{3 in}{
		\vspace*{3mm} \centering  \authorrefmark{3}Instituto Tecnologico de Orizaba\\ 
		Av. Oriente 9 No.852 \\ 
	   	94330 Orizaba, Mexico} 
}

\begin{document}

\maketitle
\thispagestyle{empty}
\pagestyle{empty}
\begin{abstract}
Human interaction introduces two main constraints: Safety and Comfort. 
Therefore service robot manipulator can't be controlled like industrial 
robotic manipulator where personnel is isolated from the robot's work envelope.
In this paper, we present a soft motion trajectory planner to try to
ensure that these constraints are satisfied. This planner
can be used on-line to establish visual and force control loop suitable in
presence of human. The cubic trajectories build by this planner are
good candidates as output of a manipulation task planner. The obtained
system is then homogeneous from task planning to robot control.

The soft motion trajectory planner limits jerk, acceleration and
velocity in cartesian space using quaternion. Experimental results
carried out on a Mitsubishi PA10-6CE arm are presented. 

\end{abstract}

\section{\textbf{INTRODUCTION}}

Arm manipulator control for industrial applications has now reached a
good level of maturity. Many solutions have been proposed for
specific utilizations. 
 However, all these applications are confined to
structured and safe spaces where no human-robot interactions occur.

Arm manipulators for human interaction need to be intrinsically safe \cite{Bicchi} \cite{Phriend}, 
but the control level has also to guarantee safety and comfort for
humans. The soft motion trajectory planner presented in this paper provides tools to 
build such systems by limiting jerk, acceleration and velocity.


The problem of robot control has been divided in two hierarchical levels; 
the lower level called \emph{control} or \emph{path tracking} and the upper level
called \emph{trajectory planning}. Using this approach, industrial robots can 
evolve at high speeds satisfying path constraints. Literature presents various
works, Geering \emph{and al} \cite{Geering} propose time-optimal  motions using a 
bang-bang control, Rajan proposes a two steps minimization algorithm \cite{Rajan}, 
temporal/torque constraints are considered in the works of Shin and 
McKay \cite{Shin}, Bobrow \emph{and al} \cite{Bobrow} and finally Kyriakopoulos 
and Saridis propose minimal jerk control \cite{Kyriakopoulos}. The objectives 
of the trajectory planner are to improve tracking accuracy and reduce 
manipulator wear by providing continuous references to the
servo-motors control. Needs for productivity improvements for
numerically controlled machine tools have generated numerous work to
optimize feed-rate. In this case path tracking accuracy
is far more important and approaches become similar. For example J. Dong \cite{Dong}
shows that limiting jerk in feed-rate optimization leads up a decrease of
contouring errors and acoustic signals.

In a human interaction context, safety is directly linked with the velocity limit and comfort with acceleration and jerk bounds. 
Such constrained movements are soft in cartesian space even for rotations, starts and stops.
This planner is used daily to plan trajectory along a path
computed by HAMP \cite{Sisbot07}, a Human Aware Motion Planner, and           
Grasp Planner  \cite{Efrain}, both using Move3D \cite{Simeon01}.
These paths are defined by lines that connect different points. The temporal evolution along the
path is then computed by the soft motion trajectory planner as
presented in the experimental section.
Beside the limitations in
jerk, acceleration and velocity provided by this approach, we hope
that this would help to integrate visual and force loop that are known
to have different time constant.


This paper presents the related work in section II. 
Section III describes the soft motion trajectory planner and section IV presents some experimental results. 


\section{\textbf{Related Work}}

%

To achieve smooth motion and tracking in task or joint space, several approaches have been presented, such as 
trapezoidal or bell-shaped velocity profiles using cubic, quartic or quintic 
polynomials.
Lloyd \cite{Lloyd} introduces a method adjusting the spatial shape of the transition curve of adjacent path segments.
Liu \cite{Liu} uses seven cubics to update on-line a smooth mono-dimensional motion.

Andersson \cite{Andersson} uses a single quintic polynomial for 
representing the entire trajectory, while Macfarlane \cite{Macfarlane} extends
Andersson's work and uses seven quintic polynomials for industrial robots.

In the case of human interaction Amirabdollahian \emph{and al} \cite{Amira} use a 
seventh order polynomial while Seki and Tadakuma \cite{Seki} propose the 
use of fifth order polynomial, both of them for the entire trajectory
with a minimum jerk model. Herrera and Sidobre \cite{Herrera} propose seven cubic
equations to obtain \emph{Soft Motions} for robot service applications.

%
%
%


\section{\textbf{Soft Motion Trajectory Planner}}
We consider the planning of a trajectory defined by a set of points generated by path planning techniques 
that the end-effector must follow in cartesian space.
We propose a soft motion trajectory planner that limits jerk, acceleration and velocity for service robot applications.

\subsection{\textbf{Monodimensional Case}}
In order to better understand elementary motions, we introduce the acceleration-velocity frame (Fig. \ref{AVFrame}). 
Then we consider the point to point canonical case of Fig. \ref{FGM}. 
Finally, we extend our approach to general cases in which initial and final kinematic conditions are not null. \\

\subsubsection{\textbf{The elementary motions in the Acceleration-Velocity frame}} 

\begin{figure}[b]
\centering 
\includegraphics[width=\columnwidth]{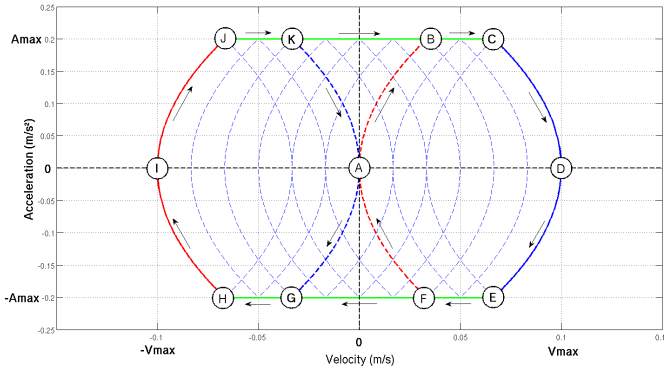} 
\caption{Acceleration-Velocity frame}
\label{AVFrame} 
\end{figure}

Initial and final conditions are defined by:
\begin{eqnarray}
  \label{IFcond}
     & A(T_0)=A_0  & A(T_f)=A_f   \nonumber \\
	 & V(T_0)=V_0  & V(T_f)=V_f   \\
     & X(T_0)=X_0  & X(T_f)=X_f  \nonumber 
\end{eqnarray}

In order to simplify the presentation, we choose :
$$J_{min}=-J_{max} \quad A_{min}=-A_{max} \quad V_{min}=-V_{max}$$
Curves $J(t)$, $A(t)$, $V(t)$, $X(t)$ respectively represent jerk, acceleration, velocity and position functions.
In the Fig. \ref{AVFrame}, the point \emph{A} corresponds to the state in which motion is stopped.
Upper line \emph{JC} and lower line  \emph{EH} respectively define maximal ($A_{max}$) and minimal ($-A_{max}$) accelerations. 
The system can stay endlessly on a point along the \emph{IAD} axis because of null acceleration.
The velocity of motion is maximal ($V_{max}$) on the point \emph{D} and minimal ($-V_{max}$) on \emph{I}. The other states are unstable states, like for example from the point \emph{C}, the only possible evolution is to join the point \emph{D}.
The \emph{CDE} parabolic curve represents an evolution at maximal jerk $J_{max}$. The \emph{HIJ} curve, in contrast, represents minimal jerk evolution ($-J_{max}$).
The acceleration axis becomes a symmetric axis of the two maximal and minimal jerk parabolas (eq. \ref{JerkParaP} \& \ref{JerkParaM}).  
 
\begin{equation}
  V(t)=V_{0} + \frac{1}{2.J_{max}} A(t)^{2} 
  \label{JerkParaP}
\end{equation}   
\begin{equation}
  V(t)=V_{0} - \frac{1}{2.J_{max}} A(t)^{2} 
  \label{JerkParaM}
\end{equation}
where $V_{0} \in [-V_{max}, V_{max}]$ is the velocity at $A(t)=0$  \\

The optimal motion is a motion with jerk, acceleration and velocity constraints successively saturated \cite{Herrera}. 
Then, we can define three elementary motions ($A_i, V_i$ and $X_i$ are initial conditions of segments) :
\begin{itemize} 
\item The motion with a \emph{saturated jerk} $\pm J_{max}$ (\emph{AB}, \emph{CD}, \emph{DE}, \emph{FA}, \emph{AG}, \emph{HI}, \emph{IJ} and \emph{KA} segments): \\
$J(t) = \pm J_{max} $ \\
$A(t) = A_i \pm  J_{max} t $  \\
$V(t) = V_i + A_i t \pm  \frac{1}{2} J_{max} t^2 $ \\ 
$X(t) = X_i + V_i t + \frac{1}{2} A_i t^2 \pm  \frac{1}{6} J_{max} t^3 $
\item The motion with a \emph{saturated acceleration} $\pm A_{max}$ (\emph{BC}, \emph{EF}, \emph{GH} and \emph{JK} segments): \\
$J(t) = 0 $ \\
$A(t) = \pm A_{max} $  \\
$V(t) = V_i \pm  A_{max} t $ \\ 
$X(t) = X_i + V_i t \pm  \frac {1}{2} A_{max} t^2 $
\item Finally, the motion with a \emph{saturated velocity} $\pm V_{max}$ (\emph{D} and \emph{I} segments): \\
$J(t) = 0 $ \\
$A(t) = 0 $ \\
$V(t) = \pm V_{max} $ \\
$X(t) = X_i \pm  V_{max} t $
\end{itemize}


\subsubsection{\textbf{The point to point motion}}  
\label{SECTION_PTPM}

In this case, initial and final conditions are defined by:
\begin{center}
$A(T_0)=0 \qquad A(T_f)= 0 $\\
$V(T_0)=0 \qquad V(T_f)= 0 $\\
$\ \ X(T_0)=0  \qquad X(T_f)=X_f$
\end{center}

\begin{figure}[H]
\centering 
\begin{center}
\includegraphics[width=\columnwidth, height=12cm]{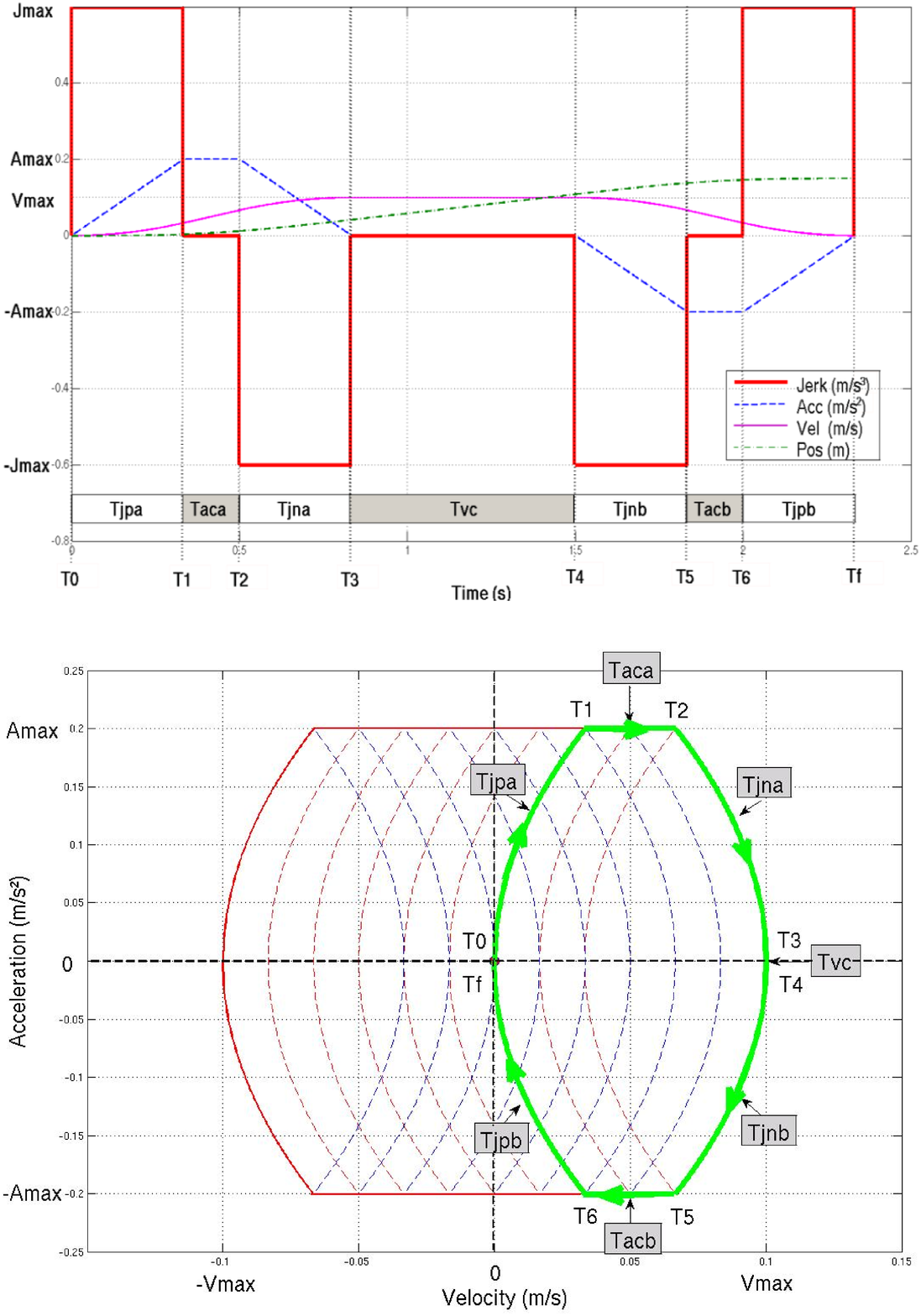} 
\caption{Jerk, Acceleration, Speed and Position curves and Motion in the Acceleration-Velocity Frame }
\label{FGM} 
\end{center}
\end{figure}

Fig. \ref{FGM} represents the optimal motion which can be separated in seven segments:\\ 
\begin{tabular}{ll}
{$T_{jpa} = T_1-T_0$}&{Jerk positive time}\\
{$T_{aca} = T_2-T_1$}&{Acceleration constant time}\\
{$T_{jna} = T_3-T_2$}&{Jerk negative time}\\
{$T_{vc}\ = T_4-T_3$}&{Velocity constant time}\\
{$T_{jnb} = T_5-T_4$}&{Jerk negative time} \\
{$T_{acb} = T_6-T_5$}&{Acceleration constant time} \\
{$T_{jpb} = T_f-T_6$}&{Jerk positive time}
\end{tabular}\\

Because of the point to point motion, it appears an anti-symmetry in acceleration and a symmetry in jerk with respect to the $T_{vc}$ segment.
Concerning the velocity curve, the symmetry effect is also present. We have then: \\
\vspace{-2mm}
$$ T_j = T_{jpa} = T_{jna} = T_{jnb} = T_{jpb} $$  
\vspace{-2mm}
$$   T_a = T_{aca} = T_{acb}  \qquad \qquad T_v = T_{vc} $$
Our system computes times $T_j$, $T_a$ and $T_v$ to get the desired soft displacement between an origin position 
and a final position. As the end effector moves under maximum motion
conditions ($J_{max}$, $A_{max}$ or $V_{max}$), we obtain a \emph{minimal time motion}. 
However, optimal motion has seven elementary motions at most as demonstrated below : \\
\emph{i)} We consider a motion composed of a constant velocity motion at $V_{max}$, which occurs during a period $dt_1$, 
and also of a constant velocity motion at $-V_{max}$ during $dt_2 \geq dt_1$. 
The motion at  $V_{max}$ balances the motion at $-V_{max}$. So, it's possible to find a motion with a shorter time which satisfies initial and final conditions. In other words, in the acceleration-velocity frame, a motion can't stay on both the $D$ and $I$ points. \\
\emph{ii)} If a motion has a constant unsaturated velocity segment, it's also possible to find a motion with a saturated velocity segment or without a constant velocity segment at all.
In both cases, the motion time is shorter. Thus, optimal motions can't have an unsaturated constant velocity segment.\\
\emph{iii)} If motion doesn't reach neither the $D$ nor $I$ points, parabolic curves can only be at the beginning or at end of motion.\\
Therefore, optimal motions can't have more than seven elementary motions.

\subsubsection{\textbf{Types of motions}}  
\begin{figure}[tp]
\includegraphics[width=\columnwidth, height=4.5cm]{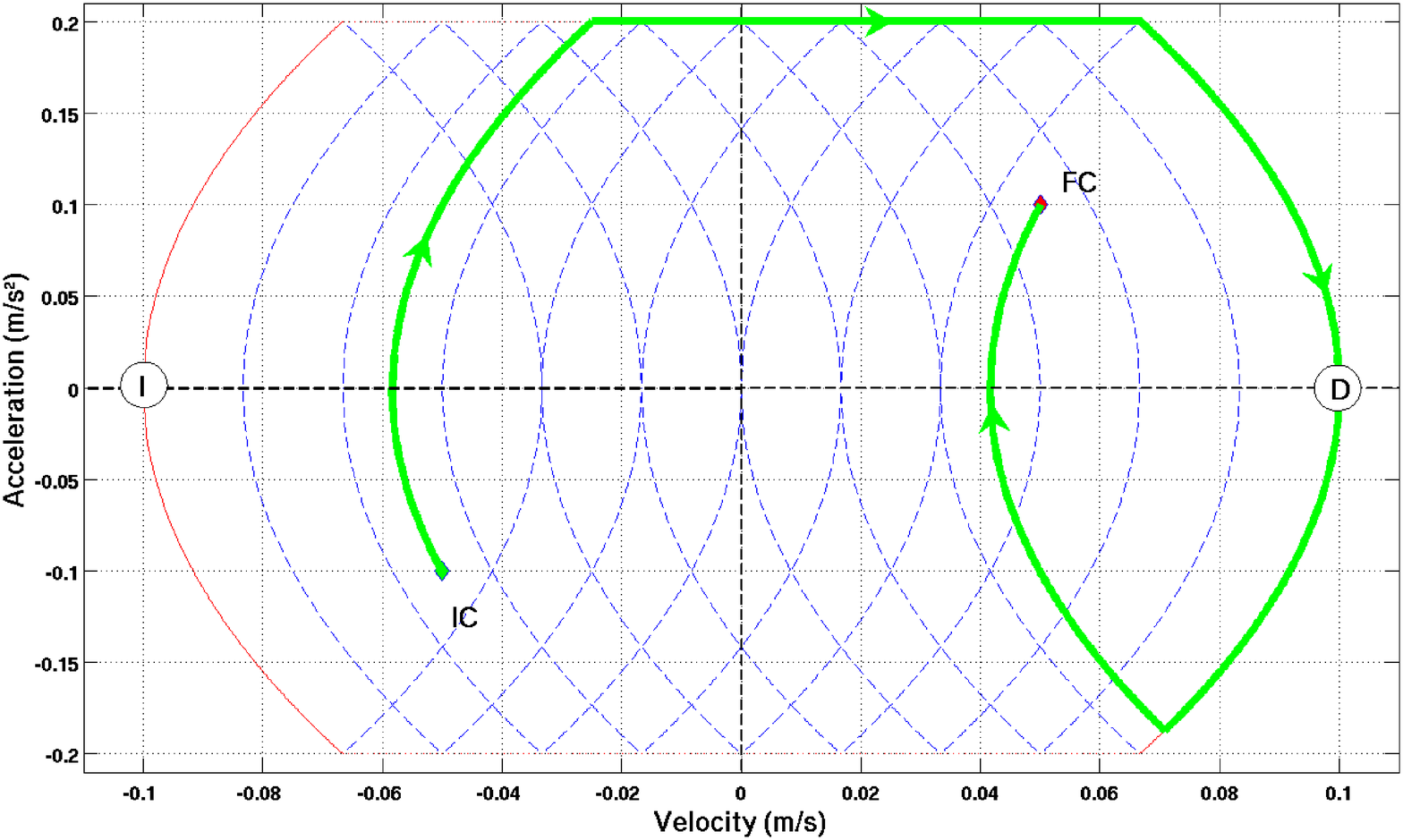} 
\includegraphics[width=\columnwidth, height=4cm]{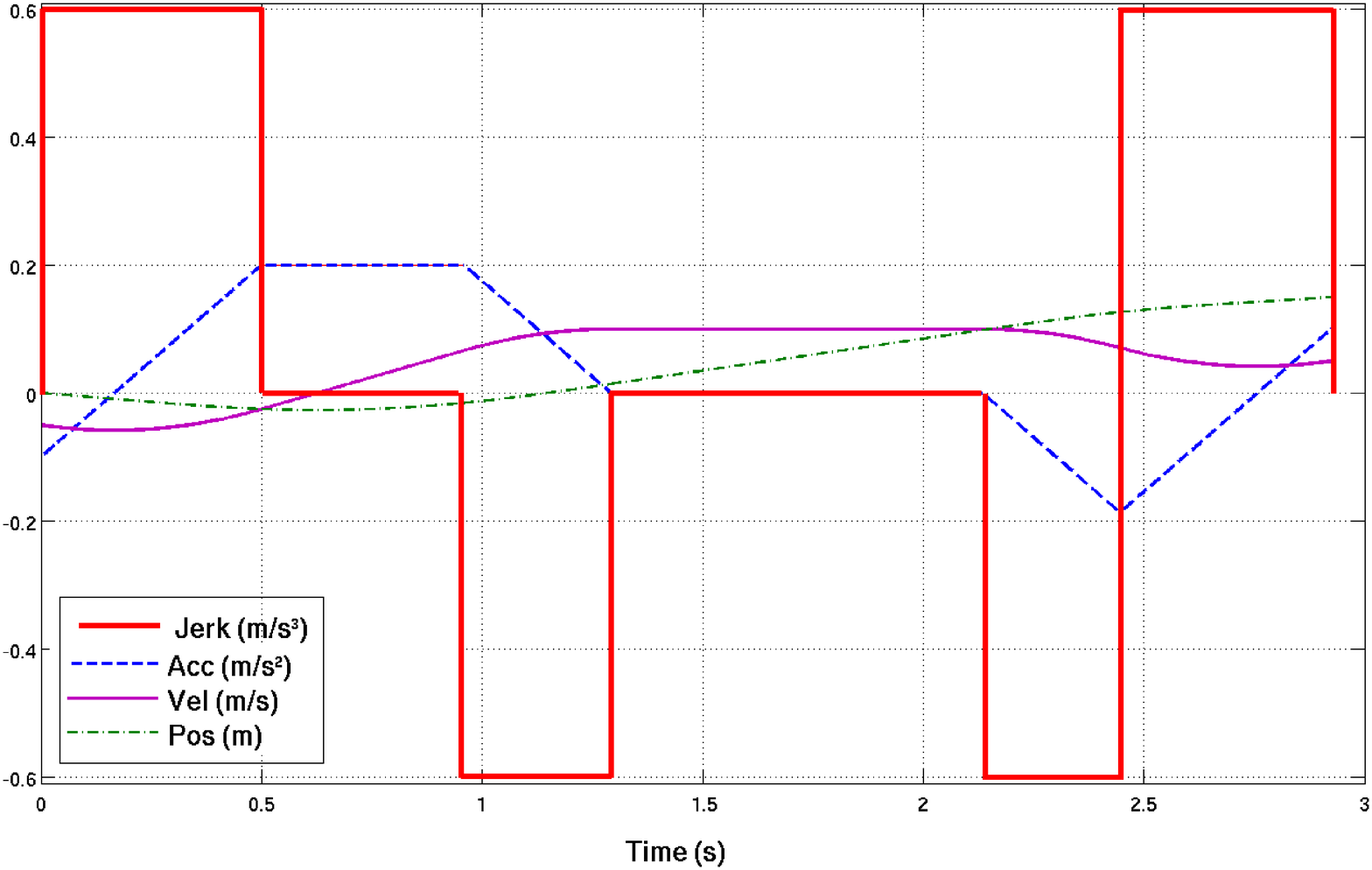} 
\caption{Motion type 1 with $Vmax$ reached}
\label{MT1} 
\includegraphics[width=\columnwidth, height=4.5cm]{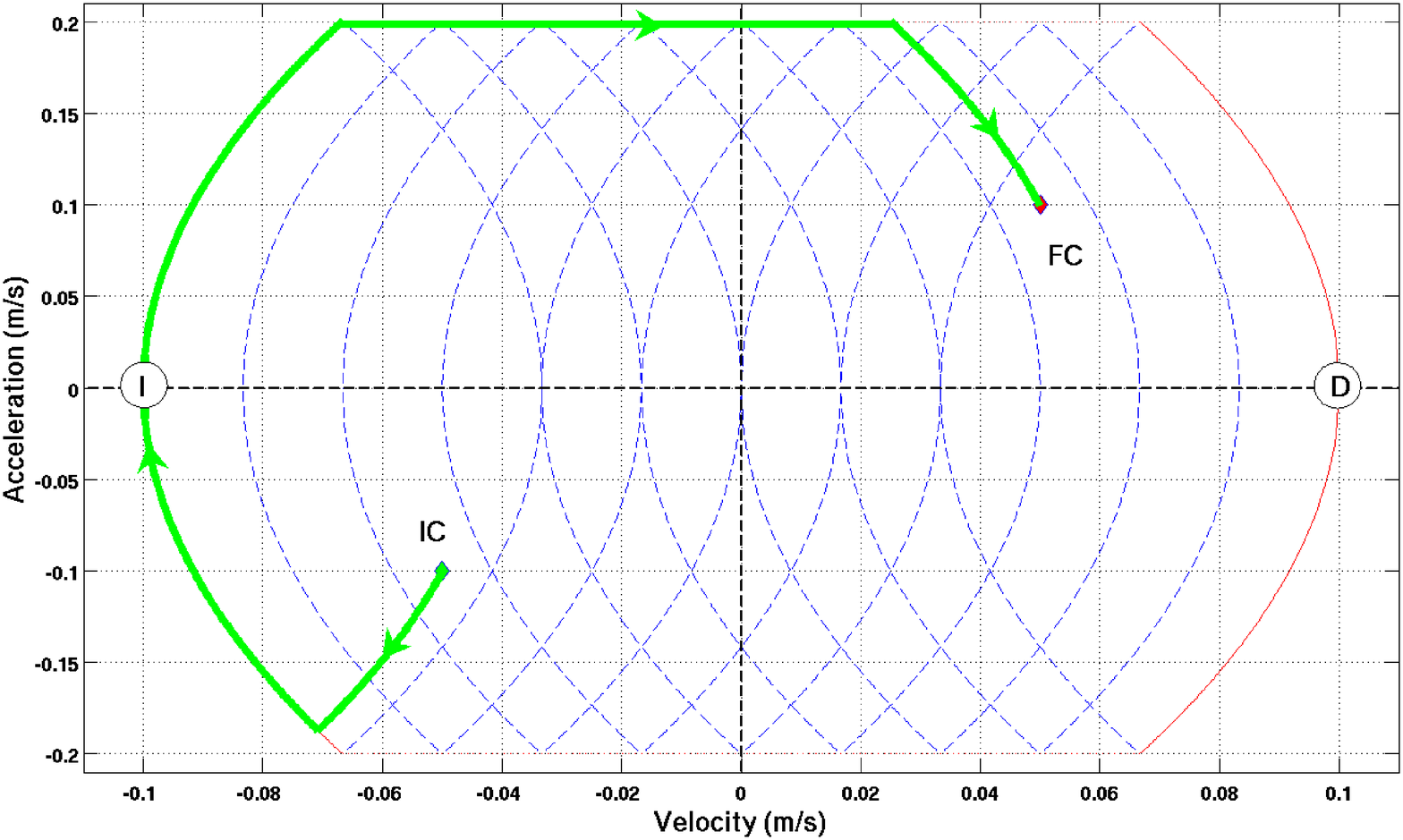} 
\includegraphics[width=\columnwidth, height=4cm]{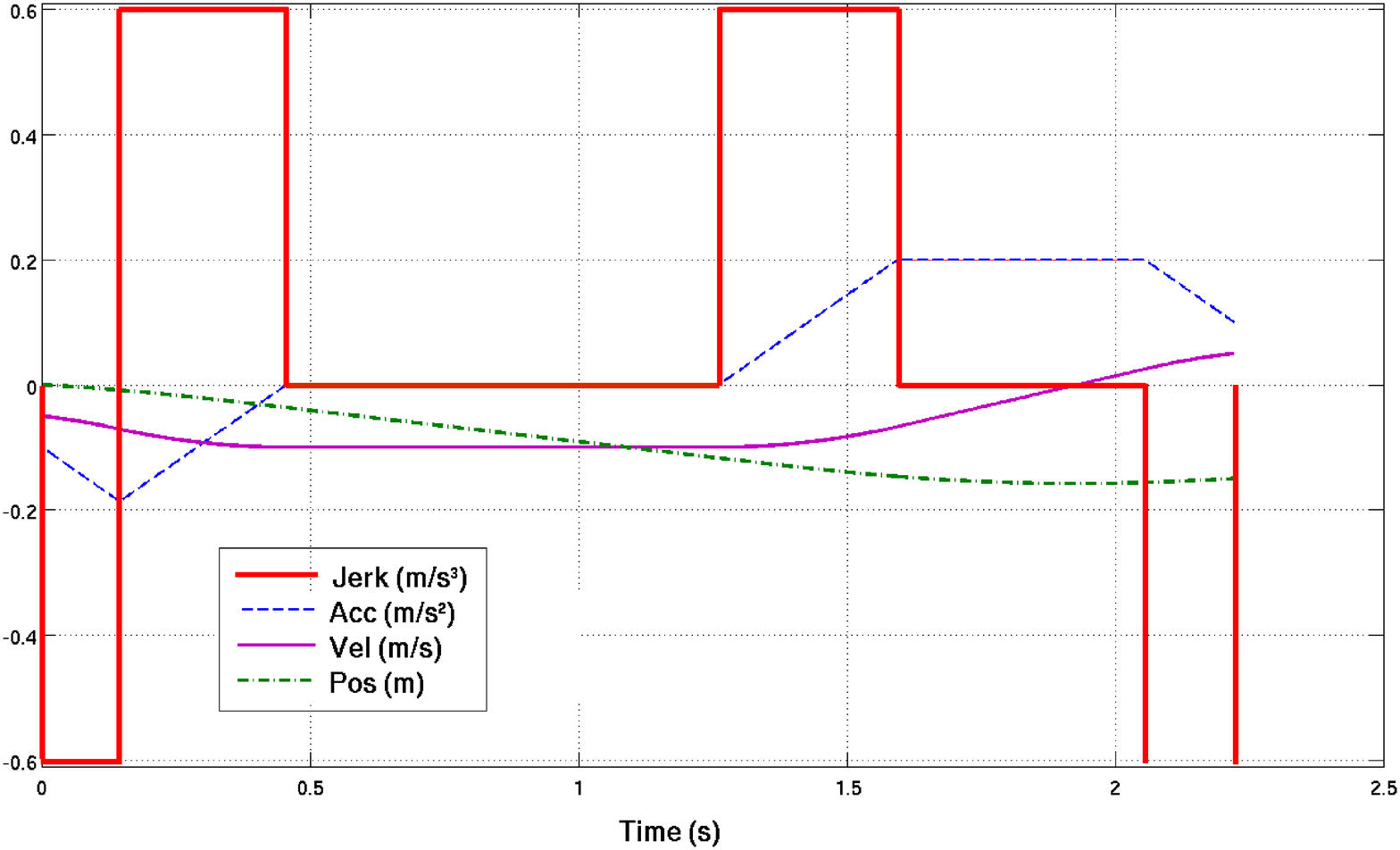} 
\caption{Motion type 2 with $-Vmax$ reached}
\label{MT2} 
\includegraphics[width=\columnwidth, height=4.5cm]{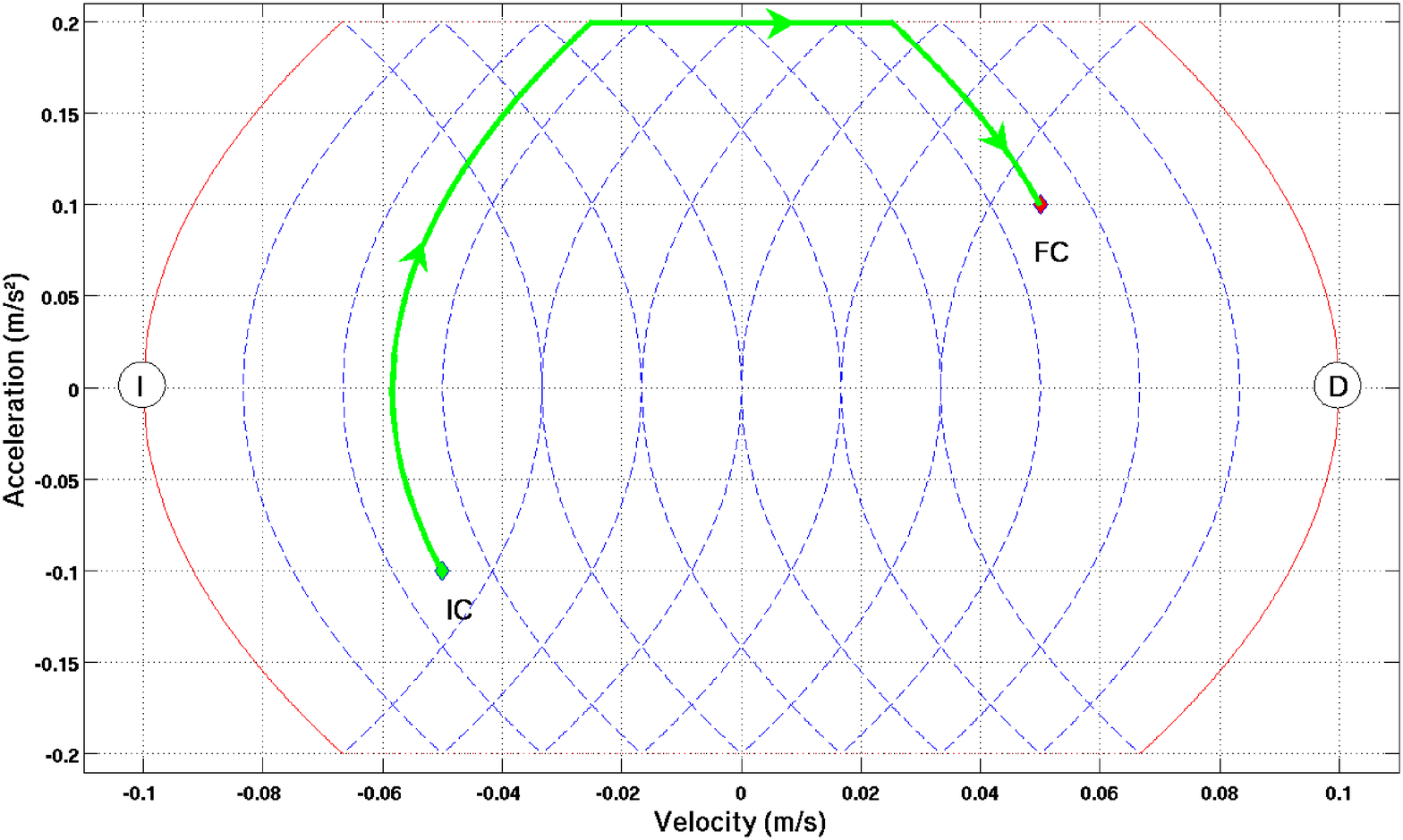} 
\caption{Motion type hybrid: \emph{Critical Motion}}
\label{MTDC} 

\end{figure}

As optimal motion is a Soft motion in minimal time, states with constant velocities 
can only be at the \emph{D} and \emph{I} points.
So, in order to attain some initial and final conditions, there are two type of motions.
A motion starting with a maximum jerk segment will be called \emph{type 1 motion} and a motion starting with a minimum jerk segment, \emph{type 2 motion}.
For example, Fig. \ref{MT1} illustrates type 1 motion which joins the point \emph{D}. Fig. \ref{MT2} illustrates type 2 motion which joins the point \emph{I}.

For short displacement, optimal motion doesn't have the constant velocity segment.  However, we have to focus on a particular motion which we call \emph{Critical motion} defined by a critical length \emph{dc}. 


\subsubsection{\textbf{The critical length}}  
\label{FTOF}
Fig. \ref{MTDC} presents critical motion that separates \emph{motion type 1} (Fig. \ref{MT1}) from \emph{motion type 2} (Fig. \ref{MT2}).
Critical length is the distance $D=X_f-X_0$ done when the time motion is minimal and separates continiously \emph{motion type 1} and \emph{motion type 2}. 
When the distance to cross becomes larger than \emph{dc}, motion is a \emph{type 1 motion}. On the other hand, when the length becomes smaller than \emph{dc}, motion is a  \emph{type 2 motion}.

\subsubsection{\textbf{The general case}}
\label{generalcase}
The previous canonical case (\ref{SECTION_PTPM}) produces simple equations because of the symmetry of curves. 
When initial and final kinematic conditions are no longer null, there is no more symmetry.
However, it's important to observe that there is an impair symmetry between type 1 motion and type 2 motion in the acceleration velocity frame. 
With this property, we can compute a type 2 motion as a type 1 motion and thus we can divide the number of algorithm's functions by two.
We have developed an algorithm which computes the time of the seven segments for type 1 motions. Because of its size, we will not detail it in this paper.
Inputs are initial and final conditions (eq. \ref{IFcond}) and the $J_{max}$, $A_{max}$ and $V_{max}$ constraints. 
This algorithm is based on thresholds that define particular lists of elementary motions. 
The most complicated cases correspond to the resolution of a six degree equation. 
This equation represents the intersection of three parabolic curves.

\subsection{\textbf{Multidimensional Case}}
We present two interesting cases of the multidimensional extension:
\begin{itemize}
\item The point to point motion: initial and final kinematic conditions are null. \label {ptpm}
\item The path following motion: the system has to pass over some points.
\end{itemize}

\subsubsection{\textbf{The point to point motion}} 
\label{ptpma} 
 Motion, in a $n$ dimensional space between two points, is a straight-line path.
The only way to ensure straight-line path is that motions have the same duration along each dimension.
To do that, we compute the final time for each dimension. 
Considering the largest motion time, we readjust the other dimension
motions to this time. 
Time adjusting is done by decreasing linearly $J_{max}$,  $A_{max}$ and  $V_{max}$.
In other words, the motion is minimum time for one direction. In the other directions, the
motions are conditioned by the minimum one. 


\subsubsection{\textbf{The path following motion}} 
\begin{figure}[bH]
\includegraphics[width=\columnwidth]{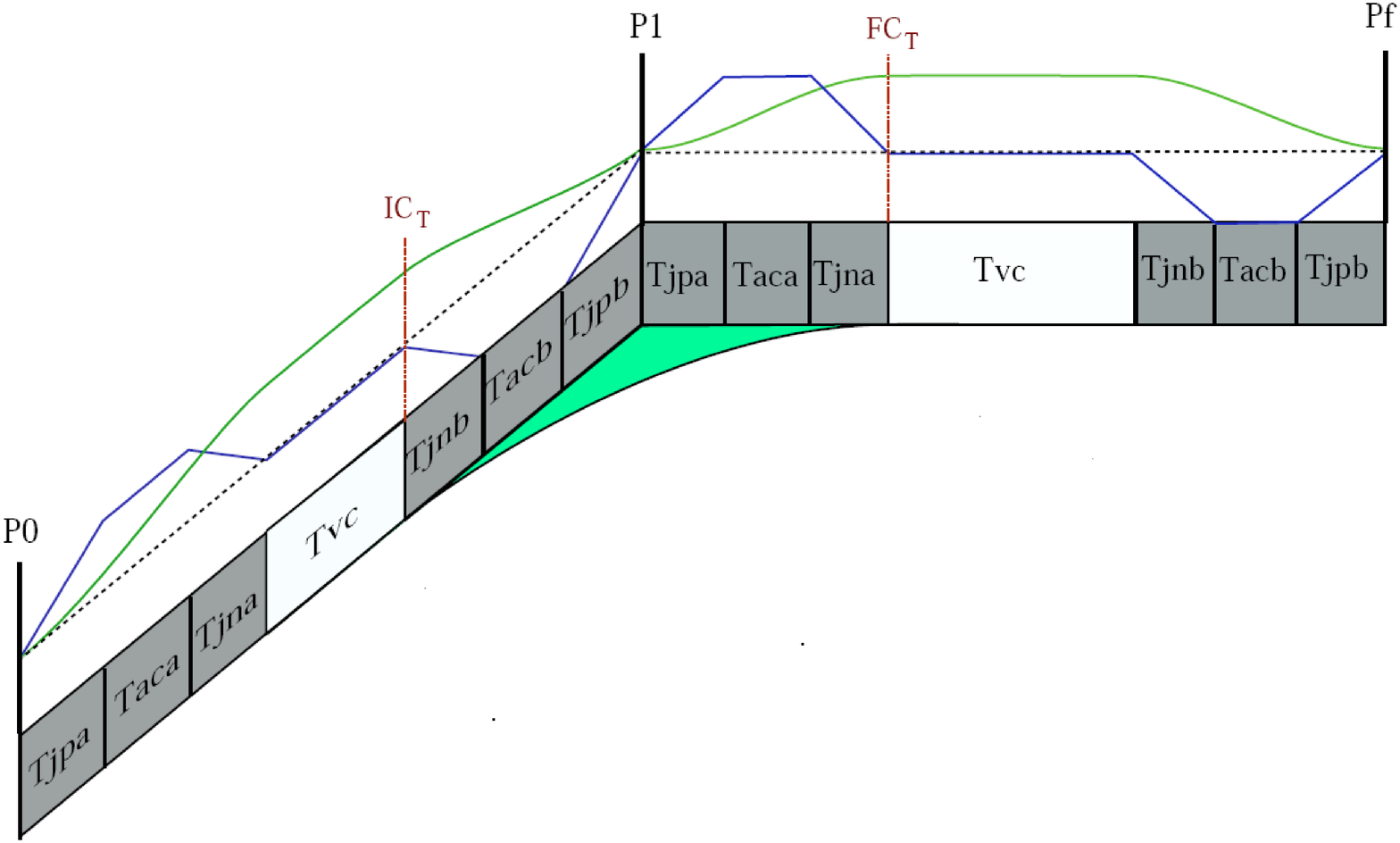} 
\caption{Planning of a motion with three points}
\label{PLANIF} 
\end{figure}

We consider a trajectory defined by points in the cartesian space (Fig. \ref{PLANIF}). At least three points are necessary: the current position of the end-effector (P0), the first target position (P1) and the final position (Pf). \\
\label{STEPS}
We describe the planification for a three points motion:\\
\emph{Step 1:} We compute the adjusted point to point motion (\ref{ptpma}) between the current position (P0) and the intermediate point (P1). We compute also the adjusted point to point motion between the point (P1) and the final point (Pf). In this state, the motion is stopped at (P1).\\
\emph{Step 2:} We use the algorithm described in \ref{generalcase} for each axis. For this transition motion, we use as initial conditions the ones found at the end point of the \emph{Tvc} segment of the first point to point motion ($IC_T$) (Fig. \ref{PLANIF}) and as final conditions the states at the beginning of the \emph{Tvc} segment of the second point to point motion ($FC_T$).
So we have for each axis :
$$A(IC_T)=  0  \qquad \ A(FC_T)=0\ \ $$
$$V(IC_T)=V_0 \qquad V(FC_T)=V_f$$
$$X(IC_T)=X_0  \qquad X(FC_T)=X_f$$
\emph{Step 3:} Once the algorithm  \ref{generalcase} is carried out, we have the optimal times $T_{opt}$ for each axis. Then, we have to constrain 
the motion time duration of each axis considering the axis which has the largest duration. We call this particular time $T_{imp}$. 

For this transition motion, we can have various type of motions like start motion, stop motion and an infinity of combinations for $V_0$ and $V_f$ velocities varying in the $[-V_{max},V_{max}]$ interval. 
The length $D=X_f-X_0$ is conditioned by $V_0$ and $V_f$. Thus, this length is a particular one because we have computed the point to point motions at the beginning. It represents the distance done when the motion is linking $V_0$ and $V_f$ passing over the point \emph{A} (Fig. \ref{AVFrame}). 
So, adjusting time duration of the transition trajectory is more difficult than the adjustment of the point to point motion. 

However, we have a particular time $T_{stop}$, the time needed to stop and restart the motion passing through (P1). If the imposed time $T_{imp}$ is larger than $T_{stop}$, we can stop the motion and adjust the duration by adding time when the motion is stopped. In the other cases when $T_{imp}$ is between $T_{opt}$ and $T_{stop}$, we have to find a combination of seven cubic segments satisfying initial and final conditions and the kinematic constraints.

Now we are breaking down different ways to adjust the duration of axis transition motions.\\
Soft transition motions must be under kinematic constraints, so we can't increase $J_{max}, A_{max}$ and $V_{max}$. Because of real time constraint, 
we don't want to solve our problem by using random or optimization algorithms.  
In \ref{ptpma}, we  adjust duration by decreasing limit conditions ($J_{max}, A_{max}$ and $V_{max}$). This strategy doesn't work anymore. Indeed,  we can't decrease $V_{max}$ in a motion if initial and final velocities are $V_{max}$. In this case, motion is only composed of a saturated velocity segment and decreasing $J_{max}$ or $A_{max}$ doesn't change the duration of the motion. When initial and final velocities are smaller than $V_{max}$, decreasing $J_{max}$ and$ A_{max}$ increase the critical length. In this way, when critical length reaches and runs over $D$, the type of motion changes and a time interval without solution appears.
So, we can't adjust motions like in \ref{ptpma}.
%

Another solution is to find a seven cubic segments with a constant velocity $V_c$ for the $T_{vc}$ segment slower than $V_{max}$ which we call \emph{Slowing Velocity Motion}. This motion isn't an optimal motion yet.
However, there are intervals with no solution if the duration of the motion vary between $T_{opt}$ and $T_{stop}$. 
Indeed, in the cases of $V_0$ and $V_f$ are near $V_{max}$, it's possible to don't have enough time to join the low velocity needed to do the distance \emph{D} in a time $T_{imp}$. The problem is that segments with saturated jerk last too long. Note that it's possible to minimize this problem by taking a jerk $J_{adj}$ bigger than $J_{max}$ in order to decrease time of jerk saturated intervals. However, increase jerk is not a good solution because motion run over kinematic constraints. 

For each axis, we compute intervals between $T_{opt}$ and $T_{stop}$ where a solution exists by computing \emph{Slowing Velocity Motion}. Then the imposed time $T_{imp}$ is the minimal time when there is a solution for each axis. An example illustrates this method in the \emph{Experimental Results} part (\ref{RESULTSTTM}).

\section{\textbf{Experimental Results}}

\subsection{\textbf{Experimental Platform}}
\begin{figure}[b]
  \centering
  \includegraphics[height=4.5cm ]{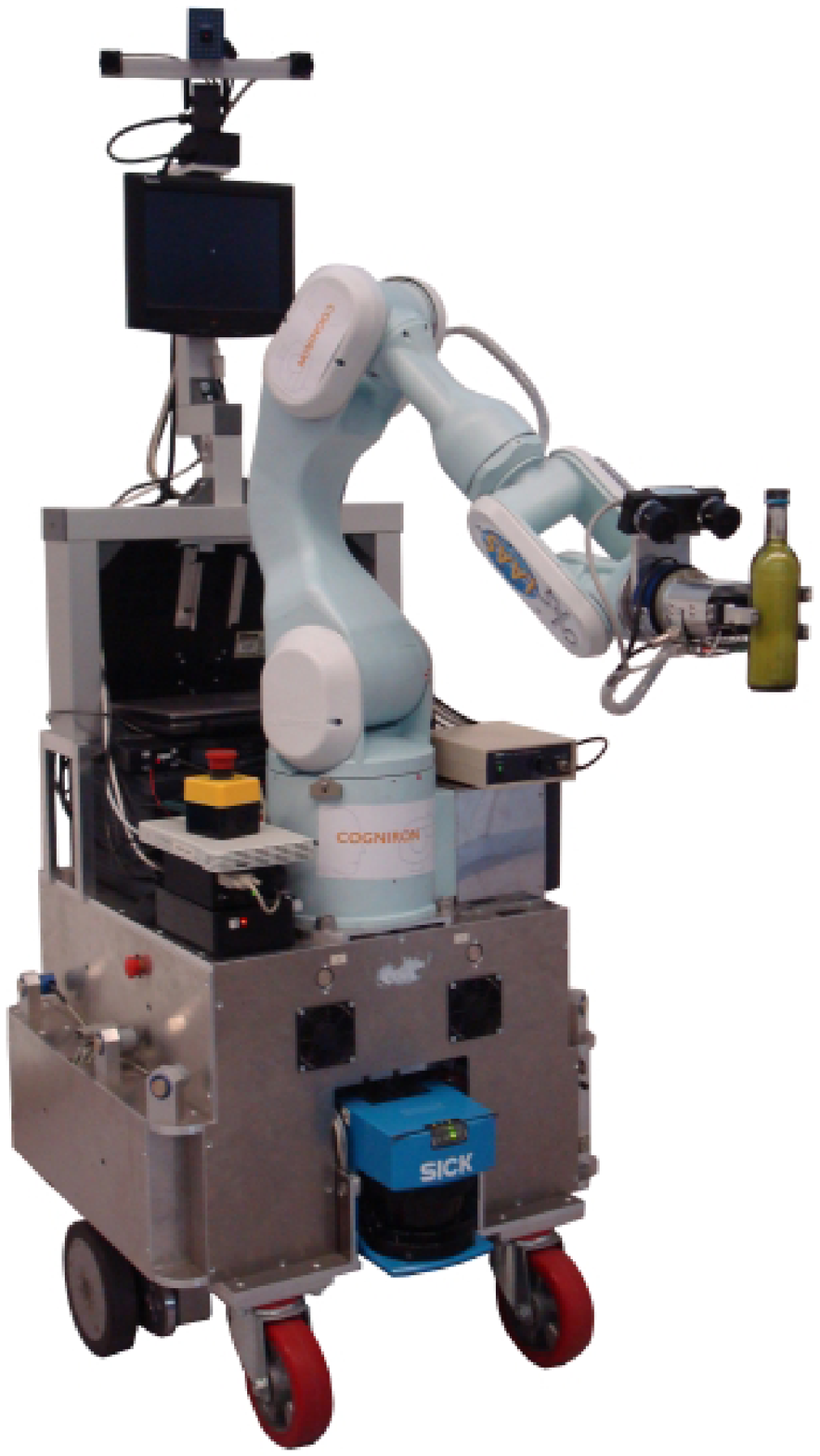}
  \caption{Our robot Jido composed of a mobile base and a 6 dof arm} 
  \label{JIDO}
\end{figure}

We implemented the soft motion trajectory planner on Jido (Fig. \ref{JIDO}), a mobile Neobotix platform MP-L655 with top mounted
manipulator PA10 from Mitsubishi.
The software control is developed using Open Robots tools: GenoM \cite{Genom}. The sampling time is fixed to 10 ms.
%
%

The linear and angular end-effector motions are limited by: 
\begin{center}
\begin{tabular}{|c|c|c|}
\hline
{ }&{Linear limits}&{Angular limits} \\
\hline
{$J_{max}$}&{$0.900 m/s^3$}&{$0.600 rad/s^3$} \\
{$A_{max}$}&{$0.300 m/s^2$}&{$0.200 rad/s^2$} \\
{$V_{max}$}&{$0.150 m/s$}&{$0.100 rad/s$} \\
\hline
\end{tabular}
\end{center}


The \emph{Pose} of the manipulator's end effector is defined by seven independent 
coordinates said \emph{Operational Coordinates}. They give the position and the
orientation of the final body in the reference frame. The advantages of using  quaternions are largely
exposed in \cite{Funda}. 

We define ${\bf P}$ for the position and ${\bf Q}$ for the orientation 
\begin{center}
$ {\bf P} = \begin {bmatrix} {x \ y \ z} \end {bmatrix} ^{T} $ \qquad  
$ {\bf Q} =\begin {bmatrix}  n  \ {\bf q} \end {bmatrix}^{T} $ \ \ 
where
\qquad $ {\bf q} = \begin {bmatrix} i   \ j  \ k \end {bmatrix} ^{T}$  
\end{center}

The linear obtained velocities $ {\bf V} $ can be directly applied as velocity references. On another hand, the evolution of the quaternion 
$ {\bf \dot Q} $ must be transformed into angular velocities. 
We use the transformation function proposed in \cite{Bruy}.
$$
\begin {bmatrix} {\bf \Omega} \\ 0
\end {bmatrix} = 2 {\bf Q_r^\top \dot Q} \qquad
\textnormal{where} \qquad
{\bf Q_r} = 
\begin {bmatrix} 
n & k & - j & i \\ 
- k & n & i & j \\ 
j & - i & n & k \\ 
- i & - j & - k & n 
\end {bmatrix}
$$

\subsection{\textbf{On-line trajectory planning without time adjustment}}
\begin{figure}[t]
  \centering
  \includegraphics[ width=\columnwidth]{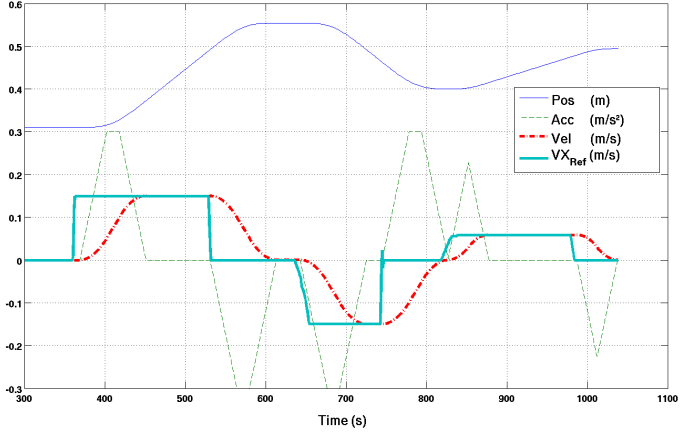}
  \caption{Soft motion of the end-effector (only X axis)} 
  \label{JOYSTICK}
\end{figure}

A 6 axis joystick (4 analog axis and 2 digital) gives velocity references ($V_{Ref}$) which must be followed by the end-effector.
$$ V_{Ref} = \begin {bmatrix} vx \ vy \ vz \ \omega x \ \omega y \ \omega z \end {bmatrix}^T$$
To track these velocities, we use the trajectory planner (\ref{generalcase}) on-line. Translation motions are independently computed.
However, we have to compute the quaternion derivative $ \dot Q$ for angular motions. As the sampling time is 10 ms, we consider that angular variation is small. So, we can use the current quaternion $Q$ as the final one to compute $ \dot Q$:\\
\begin{center}
$ \dot Q = \frac{1}{2} Q \Omega $  \qquad  with  \qquad  $ \Omega = \begin {bmatrix} \omega x \ \omega y \ \omega z \end {bmatrix}^T$ \\
\end{center}
Then, we have the vector $V_{RefPose}$:
$$ V_{RefPose} = \begin {bmatrix} vx \ vy \ vz \ \dot Q \end {bmatrix}^T$$
Trajectory is planned every 10 ms : initial conditions are the current state and final conditions are acceleration  null and velocities $V_{RefPose}$. For each direction, the distance to go $D=Xf-X0$ is the critical length for initial and final conditions. This particular length defines the shortest motion to attain final conditions. Other lengths would introduce oscillations because motion will not directly join final conditions.
Fig. \ref{JOYSTICK} illustrates the end-effector evolution for X axis.

\subsection{\textbf{Tracking trajectory motion}}
\label{RESULTSTTM}
\begin{figure}[t]
  \centering
  \includegraphics[width=\columnwidth]{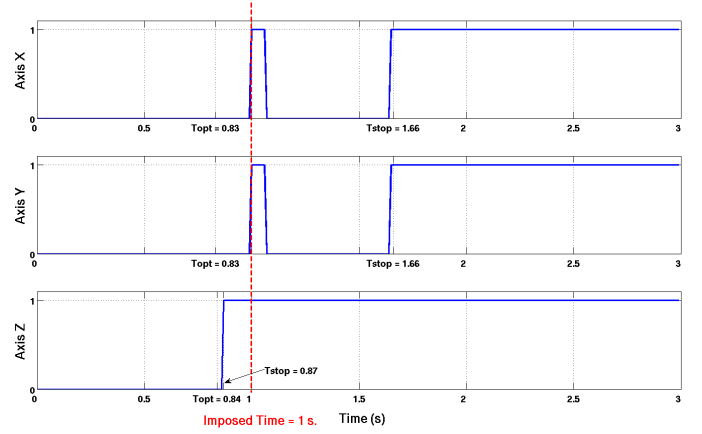}
  \caption{Time intervals where Slowing Velocity Motion works  (\textbf{0} : without solution ; \textbf{1} : with solution)} 
  \label{SUB}
\end{figure}

Even though we can do rotation, for the clarity of the presentation, we present a translation motion.
We consider the trajectory defined by the three points :
\begin{center}
$ {\bf P0} = \begin {bmatrix} X(P0)=X_0 \\Y(P0)=Y_0 \\Z(P0)=Z_0
\end {bmatrix} $ \qquad  
$ {\bf P1} =\begin {bmatrix}  X(P1)=X_0 + 0.15 \\Y(P1)=Y_0 + 0.15  \\Z(P1)=Z_0
\end {bmatrix} $ \qquad 
$ {\bf Pf} =\begin {bmatrix}  X(Pf)=X(P1) + 0.15 \\Y(Pf)=Y(P1) + 0.15  \\Z(Pf)=Z(P1) + 0.15
\end {bmatrix} $ 
\end{center}
Considering steps and notation explained on \ref{STEPS}, we compute the point to point motions between $P0$ and $P1$ and between $P1$ and $Pf$.
 So, the initial and final conditions for the transition motion are :

\begin{center}
\begin{tabular}{|c|c|c|c|}
\hline
{ }&{Axis X}&{Axis Y}&{Axis Z} \\
\hline
{V($IC_T$) (m/s)}&{$0.150$}&{$0.150$}&{$0$} \\
{V($FC_T$) (m/s)}&{$ 0.150 $}&{$0.150$}&{$0.15$} \\
{D (m)}&{$ 0.125 $}&{$0.125$}&{$0.0623 $} \\
{$T_{opt}$ (s)}&{$ 0.833 $}&{$0.833$}&{$0.84 $} \\
\hline
\end{tabular}
\end{center}
where $D$ is the axis displacement.
At this step, we have to adjust transition motion times. So, as explained in \ref{STEPS}, we compute time intervals when \emph{Slowing Velocity Motion} works. Fig. \ref{SUB} illustrates how to find $T_{imp}$ . 

More video results could be found at: \\
\emph{http://www.laas.fr/$\sim$xbroquer}
\section{\textbf{Conclusions}}

The soft motion trajectory planner presented in this paper is simpler
than previous ones and avoids the optimization stage. 
For both the point to point motion and the transition motion, series of cubic curves are computed. 
For each axis, these cubic trajectories share the same time intervals.
Due to direct computation of cubic parameters, the planner is fast enough to be used on-line. 

Experimental results show the validity of the approach for real-time control and 
trajectory planning in human presence. 
To improve task planner characteristics, we are currently incorporating the trajectory planner
into the path planner. Our objective is to directly build soft cubic
curves at the task planification level and enjoy richer families of
curves.

%
%
%
%
%





\section{\textbf{Acknowledgments}}
The research leading to these results has received funding from the
European Community´s Seventh Framework Programme (FP7/2007-2013) under
grant agreement n° 216239 with DEXMART project.


\bibliographystyle{IEEEtran.bst}
\bibliography{biblioNacho}


%
%
%

\end{document}